\documentclass{article}

\usepackage[preprint]{neurips_2025}

\usepackage[table]{xcolor}
\usepackage{colortbl}
\usepackage{multirow} 
\usepackage{tabularx}
\usepackage{url}
\usepackage{mathtools}
\usepackage{subcaption}
\usepackage{adjustbox}
\usepackage{wrapfig, capt-of}
\usepackage{multirow}
\usepackage{xspace}

\usepackage{amsmath,amsfonts,bm}

\def\eqref#1{equation~\ref{#1}}

\def\1{\bm{1}}

\DeclareMathAlphabet{\mathsfit}{\encodingdefault}{\sfdefault}{m}{sl}
\SetMathAlphabet{\mathsfit}{bold}{\encodingdefault}{\sfdefault}{bx}{n}

\newcommand*{\eg}{e.g.\@\xspace}
\newcommand*{\ie}{i.e.\@\xspace}
\newcommand{\heading}[1]{\noindent\textbf{#1.}}

\newcommand{\NICKNAME}{\textsc{STream3R}} 
\newcommand{\nickname}{\NICKNAME} 

\newcommand{\encoder}{\mathrm{Encoder}}
\newcommand{\decoder}{\mathrm{Decoder}}
\newcommand{\decoderblk}{\mathrm{DecoderBlock}}
\newcommand{\head}{\mathrm{Head}}

\newcommand{\image}{\bm{I}} 

\newcommand{\real}{\mathbb{R}}

\newcommand{\RN}[1]{%
  \textup{\uppercase\expandafter{\romannumeral#1}}%
}

\definecolor{yellow}{rgb}{1, 1, 0.7}
\definecolor{orange}{rgb}{1, 0.85, 0.7}
\definecolor{tablered}{rgb}{1, 0.7, 0.7}
\definecolor{red}{rgb}{1, 0, 0}

\newcommand{\ft}{\bm{F}_t}

\newcommand{\pmself}{\hat{\bm X}_t^\text{local}}
\newcommand{\pmcross}{\hat{\bm X}_t^\text{global}}

\newcommand{\headself}{\mathrm{Head}_\text{local}}
\newcommand{\headcross}{\mathrm{Head}_\text{global}}
\newcommand{\headpose}{\mathrm{Head}_\text{pose}}

\newcommand{\pmselfset}{\mathcal{\hat{X}}^\text{local}}
\newcommand{\pmcrossset}{\mathcal{\hat{X}}^\text{global}}
\newcommand{\pose}{\hat{\bm P}_t}

\newcommand{\duster}[0]{DUSt3R}

\newcommand{\monster}[0]{MonST3R}
\newcommand{\spanner}[0]{Spann3R} 

\usepackage[utf8]{inputenc} 
\usepackage[T1]{fontenc}    
\usepackage{url}            
\usepackage{booktabs}       
\usepackage{amsfonts}       
\usepackage{nicefrac}       
\usepackage{microtype}      
\usepackage{xcolor}         

\usepackage[colorlinks=true,
            linkcolor=cyan,
            citecolor=cyan,
            urlcolor=magenta]{hyperref}

\title{\nickname{}: Scalable Sequential 3D Reconstruction with Causal Transformer}

\author{Yushi Lan$^{1*}$,
Yihang Luo$^{1*}$,
Fangzhou Hong$^1$,
\textbf{Shangchen Zhou}$^1$, \\
\textbf{Honghua Chen}$^1$,
\textbf{Zhaoyang Lyu}$^2$,
\textbf{Shuai Yang}$^3$,
\textbf{Bo Dai}$^4$,
\textbf{Chen Change Loy}$^1$,
\textbf{Xingang Pan}$^1$
\\
$^1$S-Lab, Nanyang Technological University, Singapore \\ 
$^2$Shanghai Artificial Intelligence Laboratory
$^3$WICT, Peking University
$^4$The University of Hong Kong \\
\url{https://nirvanalan.github.io/projects/stream3r}}

\begin{document}

\maketitle
\renewcommand{\thefootnote}{\fnsymbol{footnote}}
\footnotetext[1]{Equal contribution.}

\begin{figure}[h]
  \vspace{-6mm}
    \centering
    \includegraphics[width=\textwidth]{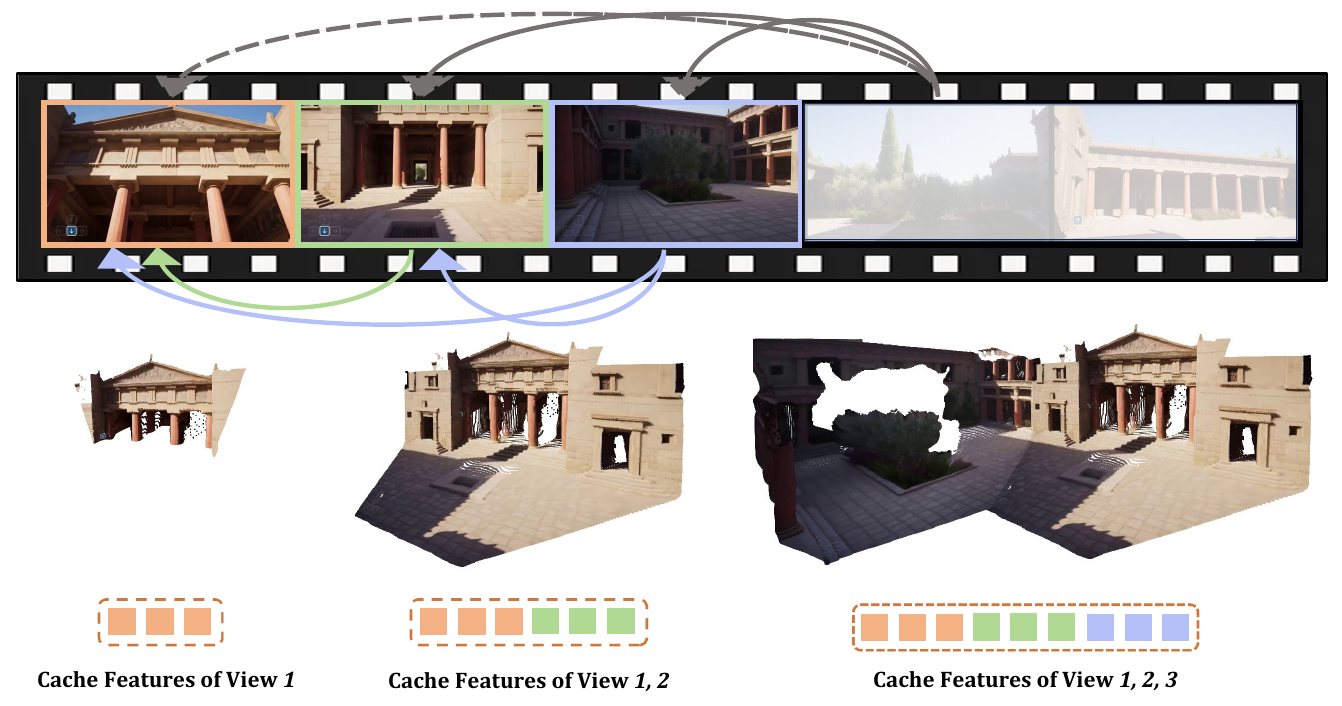}
    \caption{\nickname{}. Given a stream of input images, our method estimates dense 3D geometry for each incoming frame using a causal Transformer. Features from previously observed frames are cached as context for future inference.}
    \label{fig:teaser}
  \vspace{-2mm}
\end{figure}



\begin{abstract}
\vspace{-2mm}
We present \nickname{}, a novel approach to 3D reconstruction that reformulates pointmap prediction as a decoder-only Transformer problem. Existing state-of-the-art methods for multi-view reconstruction either depend on expensive global optimization or rely on simplistic memory mechanisms that scale poorly with sequence length. In contrast, \nickname{} introduces an streaming framework that processes image sequences efficiently using causal attention, inspired by advances in modern language modeling.
By learning geometric priors from large-scale 3D datasets, \nickname{} generalizes well to diverse and challenging scenarios, including dynamic scenes where traditional methods often fail. Extensive experiments show that our method consistently outperforms prior work across both static and dynamic scene benchmarks.
Moreover, \nickname{} is inherently compatible with LLM-style training infrastructure, enabling efficient large-scale pretraining and fine-tuning for various downstream 3D tasks. Our results underscore the potential of causal Transformer models for online 3D perception, paving the way for real-time 3D understanding in streaming environments.
More details can be found in our \href{https://nirvanalan.github.io/projects/stream3r}{project page}.

\end{abstract}

\section{Introduction}
\label{sec:intro}

Reconstructing detailed 3D geometry from images is the crux in computer vision~\cite{schonberger2016colmap,schoenberger2016mvs,chen2021mvsnerf} and serves as the pre-requisite for a series of downstream applications, like autonomous driving~\cite{kitti}, virtual reality~\cite{Zheng2023pointavatar,lan2023gaussian3diff}, robotics~\cite{irshad2024neuralfieldsroboticssurvey}, and more.
While traditional visual-geometry methods like SfM~\cite{schonberger2016colmap} and Multi-view Stereo~\cite{yao2018mvsnet,yao2019recurrent} tackle this problem by solving a series of sub-problems through handcrafted designs, a recent trend led by DUSt3R~\cite{wang2024dust3r} has demonstrated a promising new way of directly regressing point clouds using powerful transformers.  
This paradigm, along with its follow-up works including MASt3R~\cite{mast3r_arxiv24}, Fast3R~\cite{Yang_2025_Fast3R}, and VGG-T~\cite{wang2025vggt}, enables the reconstruction of 3D geometry from a number of input images--ranging from a single image to hundreds--offering a more unified solution to 3D reconstruction.

While these works focus on processing a fixed set of images, real-world applications often require continuously processing streaming visual input and updating the reconstruction on-the-fly~\cite{davison2007monoslam}, such as when an autonomous agent explores a new environment or when processing a long video sequence.
Handling streaming input poses significant new challenges.
For example, naively running Fast3R or VGG-T every time a new image arrives would incur significant redundant computation, as they have to reconstruct from scratch without inheriting previous results. 
These methods also struggle with long videos due to the expensive full-attention operation. 
Spann3R~\cite{wang2024spann3r} extends DUSt3R with a memory design~\cite{cheng2022xmem} to support incremental reconstruction, but it still suffers from significant accumulated drift and fails over dynamic scenes. 
The most relevant concurrent work is CUT3R~\cite{wang2025cut3r}, which proposes a RNN paradigm~\cite{zaremba2015recurrentneuralnetworkregularization} to handle unstructured or streaming inputs. However, the RNN-based design does not scale well with modern network architectures~\cite{dao2023flashattention2} and struggles with long-range dependency due to its limited memory size.

In light of the streaming nature of this task, in this work, we are interested in investigating \textit{the use of a transformer with uni-directional causal attention to achieve online, incremental 3D reconstruction.}
In an LLM-style transformer with causal attention, the prediction at each step reuses previous computations through a KVCache, which is proved successful in many language and audio tasks~\cite{llama,copet2024simple}. 
We observe that this property is also highly desirable for addressing online 3D reconstruciton from streaming data, as each step should build upon the previous reconstruction while integrating new content from the incoming frame.

Motivated by this, we propose \nickname{}, a comprehensive framework that performs 3D reconstruction from unstructured or streaming input images, and predicts the corresponding point maps in both world and local coordinates~\cite{Yang_2025_Fast3R}. 
Unlike concurrent works~\cite{Yang_2025_Fast3R,wang2025vggt} that resolve this issue by replacing DUSt3R's asymmetric decoders with bi-directional attention blocks~\cite{bert,videoworldsimulators2024}, \nickname{} follows the modern \emph{decoder-only}~\cite{gpt} transformer design, where incoming frames are sequentially processed and registered with causal attention~\cite{chen2025diffusionforcing}.
In this way, \nickname{} is naturally compatible with modern Large Language Models (LLMs)~\cite{llama} training and inference techniques such as window attention~\cite{jiang2023mistral7b} and KVCache~\cite{gpt}, \ie, the tokens of processed observations will be saved as reference for registering incoming frames.

%


We train our method end-to-end on a large collection of 3D data, and benchmark the proposed method on a series of downstream applications.
%
%
In summary, our key contributions are as follows:
\begin{enumerate}
\item We propose \nickname{}, a decoder-only transformer framework that reformulates dense 3D reconstruction into a sequential registration task with causal attention, enabling scalability to unstructured and streaming inputs.
\item \nickname{} is inherently compatible with modern LLM-style training and inference techniques, allowing efficient and scalable context accumulation across frames.
\item Our architecture supports both world- and local-coordinate pointmap prediction, and naturally generalizes to large-scale novel view synthesis scenarios via splatting-based rendering.
\item We train the model end-to-end on diverse 3D data and demonstrate competitive or superior performance on standard benchmarks, with strong generalization and fast inference speed.

\end{enumerate}


\section{Related Work}
\label{sec:related-works}
\vspace{-1mm}
\heading{Classic 3D Reconstruction}
Early 3D reconstruction pipelines -- such as Structure-from-Motion (SfM)~\cite{hartley2003multiple,schonberger2016colmap,tang2018ba} and SLAM~\cite{davison2007monoslam,mur2015orb,teed2021droid} -- estimate sparse geometry and camera poses from image collections via geometric reasoning. More recent approaches such as NeRF~\cite{mildenhall2020nerf,zhang2020nerf++,wang2021neus} and Gaussian Splatting~\cite{kerbl3Dgaussians,Huang2DGS2024} shift the focus to high-fidelity novel view synthesis using continuous volumetric representations. However, these methods are typically trained per-scene with no learned priors, leading to slow convergence and poor generalization to sparse or occluded inputs—a limitation sometimes referred to as the \emph{tabula rasa} assumption~\cite{wang2025cut3r}. In contrast, we adopt a data-driven approach that learns geometric priors from large-scale 3D datasets~\cite{ling2024dl3dv,reizenstein21co3d}, enabling fast and generalizable reconstruction from unstructured or streaming inputs.

\heading{Learning 3D Priors from Data}
Recent works leverage large-scale data to learn priors for depth estimation~\cite{depthanything,ke2024repurposing,hu2024-DepthCrafter}, pose+depth estimation~\cite{li2024_MegaSaM,som2024}, and bundle adjustment~\cite{wang2024vggsfm}. While these methods improve generalization, most focus on monocular depth or two-view setups, limiting their ability to reconstruct full geometry in the absence of known intrinsics~\cite{yin2023metric3d}. VGGSfM~\cite{wang2024vggsfm} introduces differentiable bundle adjustment by integrating neural feature matching with classic optimization, but remains iterative and computationally heavy, impeding scalability.
In the multi-view stereo domain, approaches such as MVSNeRF~\cite{chen2021mvsnerf,chen2024mvsplat} and MVSNet~\cite{yao2018mvsnet} integrate neural networks into the MVS pipeline but typically require known camera poses and still heavily rely on hand-crafted components to effectively incorporate 3D geometry.

\heading{Pointmap-based Representations}
Pointmap-based representations~\cite{wang2024dust3r,mast3r_arxiv24,charatan23pixelsplat,xu2024freesplatter,szymanowicz23splatter,zhang2024monst3r,gslrm2024} have recently emerged as a unifying format for dense 3D geometry prediction, aligning well with the output structure of neural networks. Compared to voxels~\cite{sitzmann_deepvoxels_2019}, meshes~\cite{gkioxari2019mesh}, or implicit fields~\cite{park2019deepsdf,mildenhall2020nerf}, pointmaps enable feedforward inference and real-time rendering, and can directly support applications such as rasterization-based rendering~\cite{kerbl3Dgaussians}, SLAM~\cite{murai2024_mast3rslam,slam3r}, and few-shot synthesis~\cite{ye2024noposplat}.
DUSt3R~\cite{wang2024dust3r} and follow-ups like MASt3R~\cite{mast3r_arxiv24} recast stereo 3D reconstruction as dense pointmap regression, jointly estimating depth, pose, and intrinsics from image pairs. However, their pairwise design fundamentally limits scalability -- requiring quadratic fusion operations and complex global alignment procedures when handling multi-view scenarios. Our approach maintains the advantages of pointmap representations while overcoming these scalability limitations.

\heading{4D Reconstruction from Monocular Videos}
Reconstructing dense geometry of dynamic scenes from monocular video is significant but challenging for conventional methods.
Recent methods~\cite{lei2024mosca,dreamscene4d,li2024_MegaSaM,kopf2021rcvd} leverages depth priors to resolve this challenge. Specifically, Robust-CVD~\cite{kopf2021rcvd} and MegaSAM~\cite{li2024_MegaSaM} requires time-consuming per-video optimization. MonST3R~\cite{zhang2024monst3r} builds on DUSt3R to output pointmaps for dynamic scenes by fine-tuning DUSt3R on the dynamic datasets. However, it still requires a sliding-window based per-video global alignment as post-processing.
In contrast, our method enables feedforward 4D reconstruction directly from monocular videos, supporting online prediction without costly per-video optimization or post-processing alignment.

\heading{Reconstruction Methods from Streaming Inputs}
Streaming approaches offer a more scalable alternative solution for the 3D reconstruction problem, represented by the monocular SLAM pipelines~\cite{davison2007monoslam,slam3r,Zhu2024NICER}.
Inspired by the existing learning-based online 3D reconstruction methods~\cite{3DR2N2,yu2021pixelnerf,Wang2021IBRNetLM},
recently Spann3R~\cite{wang2024spann3r} introduces a memory-based extension to DUSt3R, while Fast3R~\cite{Yang_2025_Fast3R} and VGG-T~\cite{wang2025vggt} replace asymmetric decoders with Transformer-based attention stacks to directly enable multi-view fusion. 
Despite these advances, these approaches still predominantly rely on global full-attention mechanisms, limiting their real-time scalability with increasing sequence length. 
CUT3R~\cite{wang2025cut3r} adopts an RNN-style architecture to process unstructured inputs incrementally, but suffers from limited memory capacity and poor compatibility with modern hardware acceleration techniques~\cite{dao2023flashattention2}. Our method fundamentally reconceptualizes pointmap prediction as a decoder-only Transformer task, enabling efficient causal inference through techniques like KVCache and windowed attention~\cite{jiang2023mistral7b,gpt}. This architectural design allows us to scale effectively to long sequences while maintaining full compatibility with modern LLM-style training infrastructure and optimization techniques, overcoming the limitations of previous approaches.

\section{Preliminaries: DUSt3R}
\label{sec:preliminary}
We reformulate DUSt3R~\cite{wang2024dust3r} to accept a stream of images as input.
In DUSt3R, each incoming image $\image_t$ is initially patchified into a set of $K$ tokens, $\ft  = \encoder(\image_t)$, where $\ft \in \real^{K \times C}$ and $\encoder$ is a weight-sharing ViT~\cite{dosovitskiy2020vit}.
Specifically, DUSt3R is designed to ingest two input images at a time, \ie, $t \in \{1,2\}$. The encoded images yield two sets of tokens:
\begin{equation}
    \bm{F}_1 = \encoder(\image_1), 
    \quad
    \bm{F}_2 = \encoder(\image_2).
\end{equation}
Afterwards, the decoder networks $\decoder_t$ reason over both of them through a series of transformer blocks with cross attention layer:
\begin{equation}
    G^i_1 = \decoderblk^i_1(G^{i-1}_1, G^{i-1}_2), 
   \quad
    G^i_2 = \decoderblk^i_2(G^{i-1}_2, G^{i-1}_1),
\end{equation}
with $i$ ranging from $1$ to $B$, representing the block index in a decoder of $B$ blocks in total.
$G^0_1 \coloneq \bm{F}_1$ and $G^0_2 \coloneq \bm{F}_2$. Finally, the corresponding regression head of each branch predicts a pointmap with an associated confidence map:
\begin{equation}
   \hat{\bm X}_{1,1}, \hat{\bm C}_{1,1}=\head_1(G^0_1, \dots, G^B_1), 
   \quad
   \hat{\bm X}_{2,1}, \hat{\bm C}_{2,1}=\head_2(G^0_2, \dots, G^B_2).
\end{equation}
Note that DUSt3R is designed for two-view inputs and requires an expensive and unscalable global alignment process to incorporate more input views.

\section{Method}
\label{sec:method}

\begin{figure}[t]
  \vspace{-3mm}
    \centering
    \includegraphics[width=\textwidth]{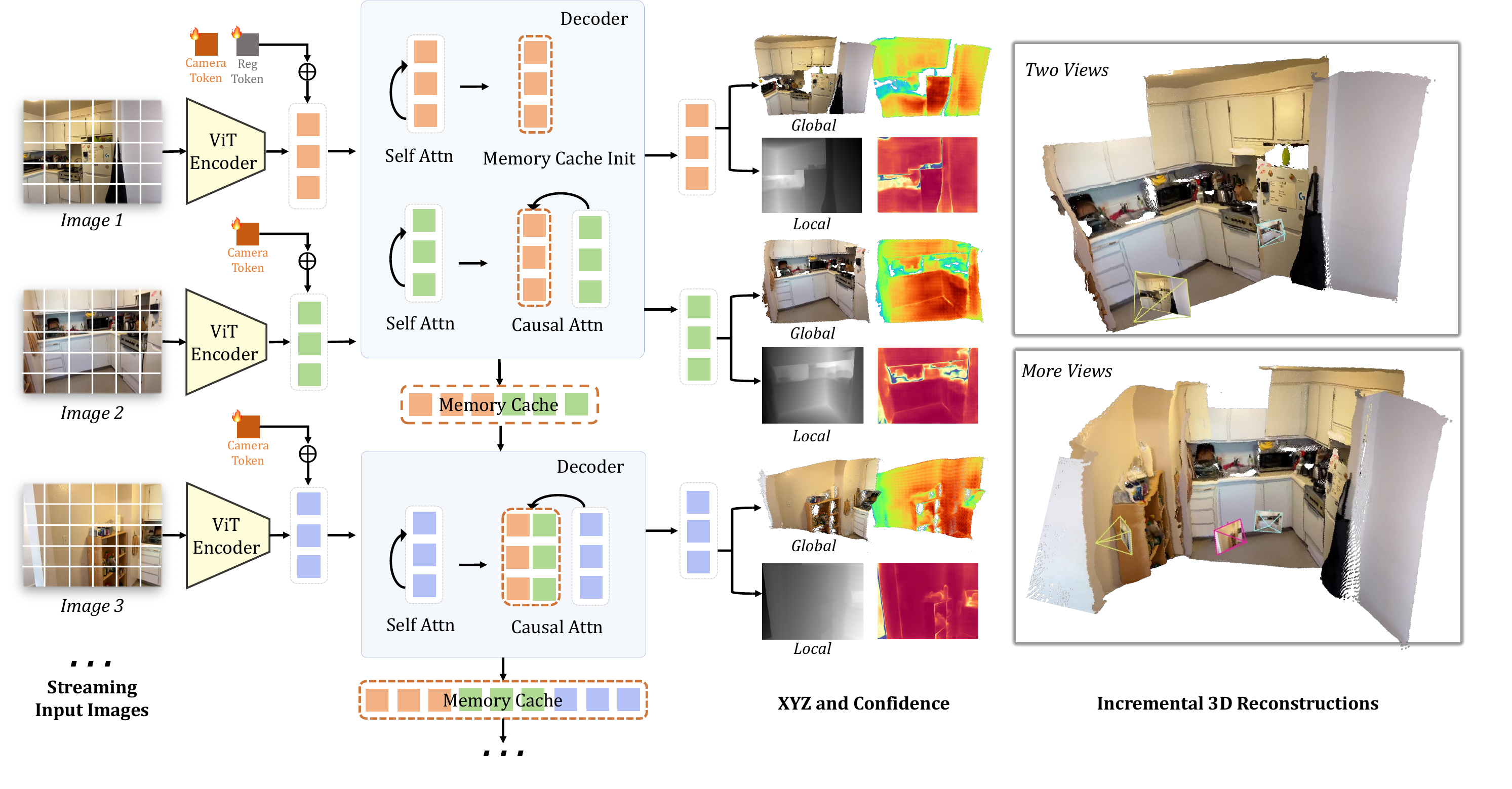}
    \caption{\textbf{Method Overview.} Built on a causal transformer, \nickname{} processes streaming images sequentially for 3D reconstruction. Each input image is first tokenized using a shared-weight ViT encoder, and the resulting tokens are passed to our causal decoder. Each decoder layer begins with frame-wise self-attention. For subsequent views, the model applies causal attention to the memory tokens cached from previous observations. The outputs include point maps and confidence maps in both world and camera coordinate systems, as long as the camera pose as shown on the right. Note that we visualize the point cloud of the $\headself$ with its depth map.}
    \label{fig:pipeline}
  \vspace{-2mm}
\end{figure}

We introduce \nickname{}, a transformer that ingests uncalibrated streaming images as inputs and yields a series of 3D attributes as output. The input can be either unstructured image collections or video. 
Unlike existing approaches~\cite{wang2025vggt,Yang_2025_Fast3R} that address this issue by adopting costly bi-directional attention over the entire input sequence or using fixed-size memory buffers~\cite{wang2024spann3r,wang2025cut3r}, 
\nickname{} instead caches the features from the past frames as \emph{context} and sequentially processes the incoming frame by performing causal attention over the accumulated observations. 
This design not only enables faster training and quicker convergence but also aligns with the architectural principles of modern LLMs, allowing us to leverage the advancement of that domain.
We first introduce the problem formulation in Sec.~\ref{method:problem_definition}, the architecture in Sec.~\ref{method:arch}, and the training objectives in Sec.~\ref{method:training}, and the implementation details in Sec.~\ref{sec:experiment}. An overview of the proposed method is shown in Fig.~\ref{fig:pipeline}.

\subsection{Problem Definition and Notation}
\label{method:problem_definition}

\nickname{} is a regression model 
that sequentially takes a streaming of $N$ RGB images $(\image)_{t}^{N}$, where each image $\image \in \real^{3\times H \times W}$ belongs to the same 3D scene.
The streaming inputs are successively transformed into a set of 3D annotations corresponding to each frame:
\begin{equation}
    f_{\theta}((\image)_{t}^{N}) = (\pmself, \pmcross, \pose)_{t}^{N}.
\end{equation}
Technically, \nickname{} is implemented as a causal transformer that maps each image $\image_t$ into its corresponding pointmap of the local coordinate $\pmself \in \real^{3\times H \times W}$ and its pointmap in a global coordinate $\pmcross \in \real^{3\times H \times W}$, which is indicated by the first input frame $\image_0$,
and its relative camera pose $\pose \in \real^9$ including both intrinsics and extrinsics. 
We devise later how these 3D attributes are predicted.

\subsection{Causal Transformer for 3D Regression}
\label{method:arch}
%

\heading{Causal Attention for Long-context 3D Reasoning}
As mentioned in Sec.~\ref{sec:preliminary}, given the streaming inputs, for each current image, $\image_t$, our method first tokenizes it into the features $\bm{F}_t = \encoder(\image_t)$. 
The main difference lies in the decoder side: rather than performing bi-directional attention over the whole sequence~\cite{Yang_2025_Fast3R} or interacting with a learnable \emph{state} as in RNN~\cite{wang2025cut3r}, we draw inspiration from the LLMs~\cite{llama,gpt,deepseekai2024deepseekv2strongeconomicalefficient} and perform causal attention efficiently with previous observations.
Specifically, after performing frame-wise self-attention in each decoder block, the current feature $G^{i-1}_t$ will cross-attend to the features of previously observed frames corresponding to the same layer:
%
%
%
\begin{equation}
    G^i_t = \decoderblk^i\left(G^{i-1}_t,\; G^{i-1}_0 \oplus G^{i-1}_1 \oplus \cdots \oplus G^{i-1}_{t-1}\right).
    \label{eq:decoderblk}
\end{equation}
This interaction ensures efficient information transfer to handle long-context dependencies.
Note that this operation is easy to implement and well optimized with KV cache during inference for efficient computation~\cite{gpt,llama}.

\heading{Simplified Decoder Design}
To achieve this, several network architecture modifications are required. 
In DUSt3R, the decoder follows a symmetric design, \ie, two separate decoders $\decoder_1$, $\decoder_2$ are employed to handle two input views. 
To extend to an arbitrary number of inputs, we remove the symmetric design and only retain a \emph{single} decoder $\decoder$ to process all the input frames.
Specifically, each block in the decoder contains a $\mathrm{SelfAttn}$ block for \emph{frame-wise} attention, and a $\mathrm{CrossAttn}$ block for causally attending to the features of all previous observations.
Note that we process the first two frames following the convention of DUSt3R due to the lack of historical context.
All incoming frames afterwards follow the causal operation in Eq.~(\ref{eq:decoderblk}).
Note that to indicate the canonical world space, we add a learnable register token $[\mathrm{reg}]$ to the tokens of the first frame $\bm{F}_1 = \bm{F}_1 + \mathrm{[reg]}$, in an element-wise manner, as shown in Fig.~\ref{fig:pipeline}. In this way, the model learns to output the global points without introducing $N$ separate decoders.
Unlike existing work~\cite{Yang_2025_Fast3R}, we did not impose positional embedding for other frames for simplicity.

\heading{Prediction Heads}
After the decoding operation, the 3D attributes corresponding to each frame can be predicted accordingly. Following existing works~\cite{Yang_2025_Fast3R,wang2025cut3r,wang2025vggt}, we predict two sets of point maps $\pmself, \pmcross$ with their corresponding confidence maps $\hat{\bm{C}}_t^{\text{local}}, \hat{\bm{C}}_t^{\text{global}}$.
Specifically, the local point map $\pmself$ is defined in the coordinate frame of the viewing camera, and the global point map $\pmcross$ is in the coordinate frame of the first image $\image_1$. 
We use two DPT~\cite{dpt} heads for point map prediction:
\begin{align}
    \pmself,\hat{\bm{C}}_t^{\text{local}} &= \headself(G^0_t, \dots, G^B_t),
    \\
    \pmcross,\hat{\bm{C}}_t^{\text{global}} &= \headcross(G^0_t, \dots, G^B_t),
    \\
    \pose &= \headpose(G^0_t, \dots, G^B_t),
\end{align}
where this redundant prediction has been demonstrated to simplify training~\cite{geo4d, Yang_2025_Fast3R} and facilitates training on 3D datasets with partial annotations, \eg, single-view depth datasets~\cite{hoi4d}.
%



\subsection{Training Objective}
\label{method:training}
\nickname{} is trained using a generalized form of the pointmap loss introduced in DUSt3R. Given a sequence of $N$ randomly sampled images, sourced either from a video or an image collection, we train the model to produce pointmap predictions denoted by $\mathcal{X} = \{\pmselfset, \pmcrossset\}$, where $\pmselfset = \{\pmself\}_{t=1}^N$ and $\pmcrossset = \{\pmcross\}_{t=1}^N$. The corresponding confidence scores are denoted as $\hat{\mathcal{C}}$.

Following DUSt3R~\cite{wang2024dust3r}, we apply a confidence-aware regression loss to the pointmaps:
\begin{equation}
    \mathcal{L}_\text{conf} = \sum_{(\hat{\bm x}, \hat{c})\in ( \mathcal{\hat{X}}, \hat{\mathcal{C}})} \left(\hat{c} \cdot \left\|\frac{\hat{\bm x}}{\hat s} - \frac{\bm x}{s}\right\|_2 - \alpha \log \hat{c} \right), 
\end{equation}
where $\hat{s}$ and $s$ are scale normalization factors for  $\mathcal{\hat{X}}$ and $\mathcal{X}$ for scale-invariant supervision~\cite{wang2024moge}. 
We also set $\hat{s}:=s$ for metric-scale datasets as in MASt3R~\cite{mast3r_arxiv24} to enable metric-scale pointmaps prediction. 
For the camera prediction loss,
we parameterize pose $\pose$ as quaternion $\hat{\bm q}_t$, translation $\hat{\bm \tau}_t$ and focal $\hat{\bm{\mathrm{f}}}_t$, and minimize the L2 norm between the prediction and ground truth:
\begin{equation}
    \mathcal{L}_\text{pose} = \sum_{t=1}^N
    \left( 
    \left\|\hat{\bm q}_t - \bm q_t\right\|_2 
    + \left\| \frac{\hat{\bm \tau}_t}{\hat{s}} - \frac{{\bm \tau}_t}{s} \right\|_2 
    +\left\|\hat{\bm f}_t - \bm f_t\right\|_2 
    \right).
\end{equation}


\section{Experiments} 
\label{sec:experiment}


\begin{figure}[t]
  \vspace{-4mm}
    \centering
    \includegraphics[width=\textwidth]{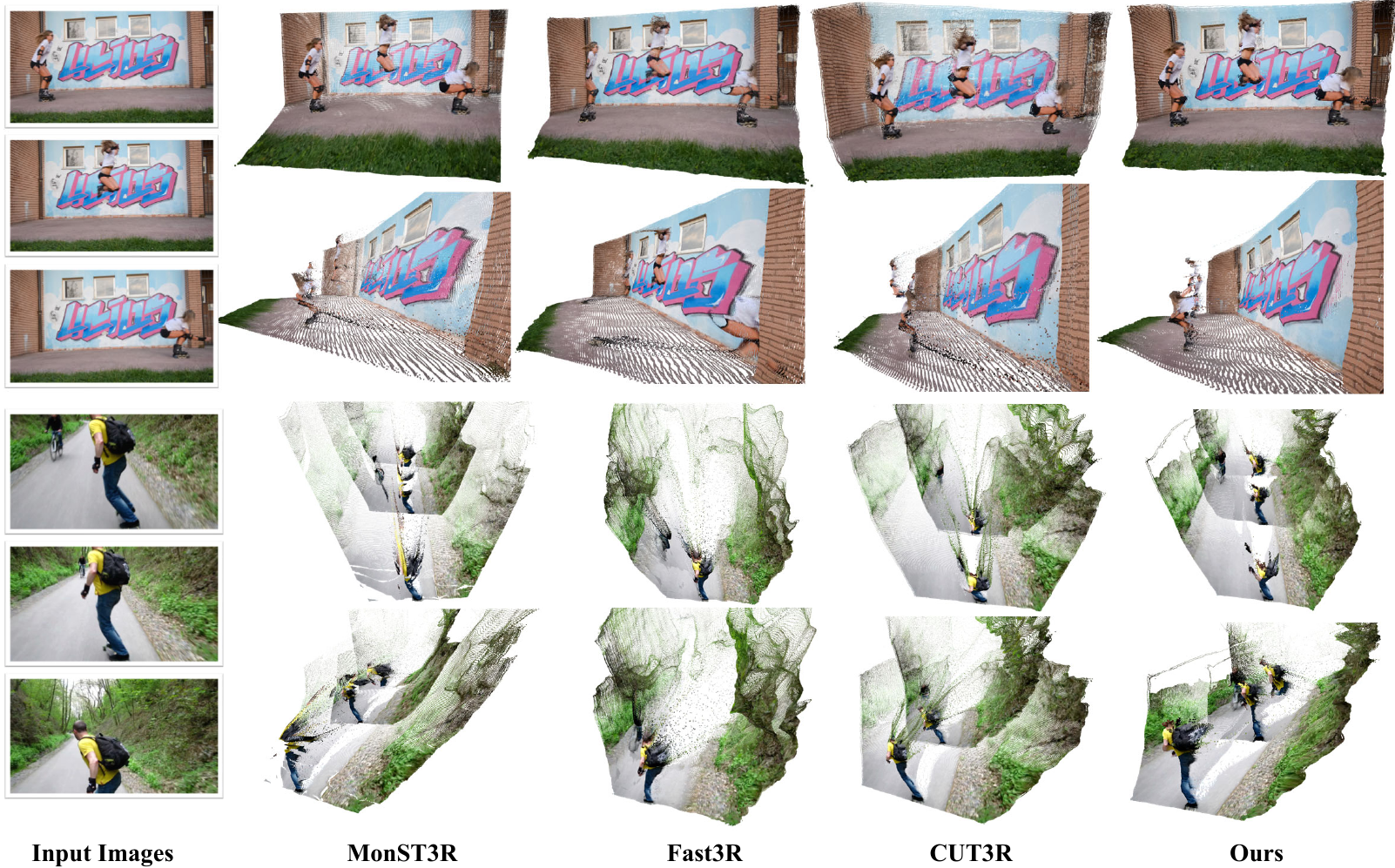}
    \caption{\textbf{Qualitative results on in-the-wild Images.} We compare our method with approaches MonST3R, Fast3R, and CUT3R, and demonstrate that it achieves superior visual quality.}
    \label{fig:qualitive_comparisons}
  \vspace{-2mm}
\end{figure}

\heading{Datasets} 
We train our method on a large and diverse collection of 3D datasets, \eg, Co3Dv2~\cite{reizenstein21co3d}, 
ScanNet++~\cite{yeshwanth2023scannet++},
ScanNet~\cite{dai2017scannet}, HyperSim~\cite{hypersim}, Dynamic Replica~\cite{karaev2023dynamicstereo}, DL3DV~\cite{ling2024dl3dv}, BlendedMVS~\cite{yao2020blendedmvs}, Aria Synthetic Environments~\cite{Pan_2023_ariadigitaltwin}, TartanAir~\cite{tartanair2020iros}, MapFree~\cite{arnold2022mapfree}, MegaDepth~\cite{Li2018MegaDepthLS}, and ARKitScenes~\cite{dehghan2021arkitscenes}.
Please check the supplement for the full dataset details.

\heading{Implementation Details}
We provide two version of \nickname{}, where \nickname{}$^\alpha$ is inspired and fine-tuned from DUSt3R~\cite{wang2024dust3r} pre-trained weights, and \nickname{}$^\beta$ is initialized from the flagship VGG-T~\cite{wang2025vggt} model. 
For \nickname{}$^\alpha$, we inherit the 24-layer CroCo ViT~\cite{croco,croco_v2} as our encoder, and retrofit its 12-layer decoder network by only retaining the first decoder $\decoder = \decoder_1$. The DPT-L~\cite{dpt} heads are used to map the decoded tokens to the local and global point maps accordingly. 
For \nickname{}$^\beta$, we replace the $\mathrm{SelfAttn}$ layer in the Global Attention of VGG-T with $\mathrm{CausalAttn}$ and fine-tune all the parameters. For memory-efficient and stable training, we inject QK-Norm~\cite{megavit} to each transformer layer and leverage FlashAttention~\cite{dao2023flashattention2} for BFloat16 mixed precision training.
%
%

\heading{Training Details}
Our model is trained with the AdamW optimizer on a batch size of $64$ with a learning rate 1e-4 for $400K$ iterations. For each batch, we randomly sample $4-10$ frames from a random training scene. The input frames are cropped into diverse resolutions, ranging from $224\times 224$ to $512 \times 384$ to improve generality.
The training runs end-to-end on $8$ NVIDIA A100 GPUs over seven days.
Gradient checkpointing is also adopted to optimize memory usage.

\heading{Baselines}
We compare our methods against a set of baselines that are designed to take a pair of views as input: DUSt3R~\cite{wang2024dust3r}, MASt3R~\cite{mast3r_arxiv24}, and MonST3R~\cite{zhang2024monst3r}. 
Besides, we include the comparison against concurrent methods
Spann3R~\cite{wang2024spann3r}, CUT3R~\cite{wang2025cut3r}, SLAM3R~\cite{slam3r}, and Fast3R~\cite{Yang_2025_Fast3R} 
that are specifically designed for handling a varying number of input images. 
We also include the flagship 3D geometry model VGG-T~\cite{wang2025vggt} for reference.
Note that Fast3R and VGG-T are bi-directional attention methods, and we group them together with methods that require global optimization (GA). We group other concurrent methods together as streaming methods that supports processing sequential inputs.
%
Note that for all methods except for VGG-T and \nickname{}$^\beta$, we conduct inference with the largest dimension of $512$. For VGG-T based methods, we conduct inference with the largest dimension of $518$ due to the requirement of DINO-V2 tokenizer~\cite{oquab2023dinov2}.
Regarding FPS, we benchmark the inference speed on the A100 GPU with FP32.
Comparisons of more concurrent methods~\cite{streamVGGT,depthanything} are included in the supplement.
%



\subsection{Monocular and Video Depth Estimation}

\begin{table}[t]
\centering
\caption{\textbf{Single-frame Depth Evaluation.} We report the performance on Sintel, Bonn, KITTI, and NYU-v2 (static) datasets. 
The best and second best results in each category are \textbf{bold} and \underline{underlined} respectively.
Our method achieves better or comparable performance against existing methods.
}
\vspace{4mm}
\footnotesize
\renewcommand{\arraystretch}{0.8}
\renewcommand{\tabcolsep}{2pt}
\resizebox{0.85\linewidth}{!}{
\begin{tabular}{@{}>{\centering\arraybackslash}m{1.4cm}
                >{\centering\arraybackslash}c >{\centering\arraybackslash}c|
                >{\centering\arraybackslash}c >{\centering\arraybackslash}c|
                >{\centering\arraybackslash}c >{\centering\arraybackslash}c|
                >{\centering\arraybackslash}c >{\centering\arraybackslash}c@{}}
\toprule
\multirow{3}{*}{\textbf{Method}} 
  & \multicolumn{2}{c}{\textbf{Sintel}} 
  & \multicolumn{2}{c}{\textbf{Bonn}} 
  & \multicolumn{2}{c}{\textbf{KITTI}} 
  & \multicolumn{2}{c}{\textbf{NYU-v2}} \\
\cmidrule(lr){2-3} \cmidrule(lr){4-5} \cmidrule(lr){6-7} \cmidrule(lr){8-9}
  & {\footnotesize Abs Rel $\downarrow$} & {\footnotesize $\delta$\textless{}$1.25\uparrow$} 
  & {\footnotesize Abs Rel $\downarrow$} & {\footnotesize $\delta$\textless{}$1.25\uparrow$} 
  & {\footnotesize Abs Rel $\downarrow$} & {\footnotesize $\delta$\textless{}$1.25\uparrow$} 
  & {\footnotesize Abs Rel $\downarrow$} & {\footnotesize $\delta$\textless{}$1.25\uparrow$} \\
\midrule
VGG-T
& \bf{0.271} & \bf{67.7} & \textbf{0.053} & \textbf{97.3} & \bf{0.076} & \underline{93.3} & \bf{0.060} & \bf{94.8} \\ 
Fast3R & {0.502} & {52.8} & {0.192} & {77.3} & {0.129} & {81.2} & {0.099} & {88.9} \\ 
DUSt3R & 0.424 & {58.7} & {0.141} & 82.5 & 0.112 & 86.3 & \underline{0.080} & \underline{90.7} \\ 
MASt3R & \underline{0.340} & \underline{60.4} & {0.142} & {82.0} & \underline{0.079} & \bf{94.7} & {0.129} & 84.9 \\ 
MonST3R & {0.358} & 54.8 & \underline{0.076} & \underline{93.9} & {0.100} & {89.3} & 0.102 & 88.0 \\ 
\midrule
Spann3R & {0.470} & 53.9 & {0.118} & {85.9} & {0.128} & {84.6} & 0.122 & 84.9 \\ 
%
%
CUT3R & {0.428} & {55.4} & \underline{0.063} & \underline{96.2} & {0.092} & \underline{91.3} & \underline{0.086} & \underline{90.9} \\ 
\nickname{}$^{\alpha}$  & \underline{0.350} & \underline{59.0} & {0.075} & {93.4} & \underline{0.088} & \underline{91.3} & {0.091} & {89.9} \\ 
\nickname{}$^{\beta}$ & \textbf{0.228} & \textbf{70.7} & \textbf{0.061} & \bf{96.7} & \textbf{0.063} & \textbf{95.5} & \bf{0.057} & \bf{95.7} \\ 
\bottomrule
\end{tabular}
}
\vspace{-4mm}
\label{tab:single_frame_depth}
\end{table}

\heading{Mono-Depth Estimation}
Following previous methods~\cite{zhang2024monst3r,wang2025cut3r}, we first evaluate monocular depth estimation on Sintel~\cite{sintel}, Bonn~\cite{Bonn}, KITTI~\cite{kitti}, and NYU-v2~\cite{nyuv2} datasets, which cover dynamic and static, indoor and outdoor, realistic and synthetic data. These datasets are not used for training and are suitable for benchmarking the zero-shot performance across different domains.
Our evaluation includes the absolute relative error (Abs Rel) and percentage of inlier points within a 1.25-factor of true depth $\delta < 1.25$, following the convention of existing methods~\cite{hu2024-DepthCrafter,depth_anything_v1}.
Per-frame median scaling is imposed as in DUSt3R. We include the quantitative results in Tab.~\ref{tab:single_frame_depth}. As can be seen, our method achieves state-of-the-art compared to streaming-based methods, and even performs best compared to VGG-T on Sintel, KITTI, and NYU-2. Also note that our method uses fewer datasets and compute resources compared to CUT3R. Specifically, CUT3R adopts a curriculum training of four stages for $100+35+40+10=185$ epochs, while our method is trained end-to-end for only $7$ epochs using a partial of CUT3R's datasets due to the computational resources constraints.

\begin{table*}[t]
  \vspace{4mm}
\centering
\caption{{\textbf{Video Depth Evaluation.}
We evaluate scale-invariant and metric depth accuracy on the Sintel, Bonn, and KITTI datasets. Methods that require global alignment are denoted as "GA". 
The "Type" column indicates whether the method is Optimzation-based ("Optim), streaming ("Stream"), or full-attention ("FA)
We also report inference speed in FPS on the KITTI dataset using 512$\times$144 resolution for all methods on an A100 GPU, except for \spanner{}, which supports Stream 224$\times$224 inputs.
Our method achieves performance that is better than CUT3R, while offering FAter inference.
For \nickname{}$^{\beta}$-W[5], we indicate using sliding window attention on \nickname{}$^{\beta}$ with window size 5.
Note that \nickname{}$^{\beta}$-W[5] achieves the fastest FPS among all streaming-based methods.
}
}
\renewcommand{\arraystretch}{1.02}
\renewcommand{\tabcolsep}{1.5pt}
\resizebox{0.99\textwidth}{!}{
\begin{tabular}{@{}ccc>{\centering\arraybackslash}p{1.5cm}>{\centering\arraybackslash}p{1.5cm}|>{\centering\arraybackslash}p{1.5cm}>{\centering\arraybackslash}p{1.5cm}|>{\centering\arraybackslash}p{1.5cm}>{\centering\arraybackslash}p{1.5cm}|>{\centering\arraybackslash}p{1.2cm}@{}}
\toprule
 &  &  & \multicolumn{2}{c}{\textbf{Sintel}} & \multicolumn{2}{c}{\textbf{Bonn}} & \multicolumn{2}{c}{\textbf{KITTI}} & \\ 
\cmidrule(lr){4-5} \cmidrule(lr){6-7} \cmidrule(lr){8-9}
\textbf{Alignment} & \textbf{Method} & \textbf{Type} & {Abs Rel $\downarrow$} & {$\delta$\textless $1.25\uparrow$} & {Abs Rel $\downarrow$} & {$\delta$\textless $1.25\uparrow$} & {Abs Rel $\downarrow$} & {$\delta$ \textless $1.25\uparrow$} & \textbf{FPS} \\ 
\midrule

& VGG-T~\cite{wang2025vggt} & FA & \bf{0.297} & \bf{68.8} & \textbf{0.055} & \textbf{97.1} & \textbf{0.073} & \textbf{96.5} & \underline{7.32} \\
& Fast3R~\cite{Yang_2025_Fast3R} & FA & 0.653 & {44.9} & {0.193} & {77.5} & \underline{0.140} & \underline{83.4} & \bf 47.23 \\

\multirow{6}{*}{\begin{minipage}{3cm}Per-sequence scale\end{minipage}} 
& DUSt3R-GA~\cite{wang2024dust3r} & Optim & 0.656 & {45.2} & {0.155} & {83.3} & {0.144} & {81.3} & 0.76 \\
& MASt3R-GA~\cite{mast3r_arxiv24} & Optim & 0.641 & {43.9} & {0.252} & {70.1} & {0.183} & {74.5} & 0.31 \\
& MonST3R-GA~\cite{zhang2024monst3r} & Optim & \underline{0.378} & \underline{55.8} & \underline{0.067} & \underline{96.3} & {0.168} & {74.4} & 0.35 \\
\cmidrule(lr){2-10}
& Spann3R~\cite{wang2024spann3r} & Stream & 0.622 & {42.6} & {0.144} & {81.3} & {0.198} & {73.7} & 13.55 \\
& CUT3R~\cite{wang2025cut3r} & Stream & {0.421}  & {47.9} & {0.078} & {93.7} & {0.118} & {88.1} & 16.58 \\
& \nickname{}$^{\alpha}$ & Stream & {0.478}  & {51.1} & {0.075} & {94.1} & {0.116} & {89.6} &  \underline{23.48} \\
& \nickname{}$^{\beta}$ & Stream & \textbf{0.264}  & \textbf{70.5} & \underline{0.069} & \underline{95.2} & \bf{0.080} & \underline{94.7} &  12.95 \\
& \nickname{}$^{\beta}$-W[5] & Stream & \underline{0.279}  & \underline{68.6} & \bf{0.064} & \bf{96.7} & \underline{0.083} & \bf{95.2} &  \textbf{32.93} \\

\midrule
\multirow{3}{*}{\begin{minipage}{3cm}Metric scale\end{minipage}} 
& MASt3R-GA~\cite{mast3r_arxiv24} & Optim & \textbf{1.022}  & 14.3 & 0.272 & 70.6 & 0.467 & 15.2  & 0.31 \\  
& CUT3R & Stream & \underline{1.029} & \textbf{23.8} & \underline{0.103} & \underline{88.5} & \textbf{0.122} & \textbf{85.5}  &  \underline{16.58} \\
& \nickname{$^{\alpha}$} & Stream & {1.041} & \underline{21.0} & \textbf{0.084} & \textbf{94.4} & \underline{0.234} & \underline{57.6}  &  \textbf{23.48} \\
\bottomrule
\end{tabular}
}
\label{tab:video_depth}
\end{table*}

\heading{Video Depth Estimation}
We also benchmark our model on the video depth task, which evaluates both per-frame depth quality and inter-frame depth consistency by aligning the output depth maps to the ground truth depth maps using a given per-sequence scale. Metric point map methods like MASt3R, CUT3R, and ours are also reported without alignment. The quantitative results for both methods are included in Tab.~\ref{tab:abla:video_depth}. Over per-sequence scale alignment, our method surpasses optimization-based baselines DUSt3R-GA~\cite{wang2024dust3r} and MASt3R-GA~\cite{mast3r_arxiv24} (static-scene assumption) and even MonST3R-GA~\cite{zhang2024monst3r} (dynamic-scene, optical flow~\cite{teed2020raftrecurrentallpairsfield} dependent).
Against the streaming state-of-the-art CUT3R, we achieve higher accuracy on all three benchmarks while running $40\%$ faster.
\nickname{} also outperforms full-attention Fast3R~\cite{Yang_2025_Fast3R}, streaming approaches Spann3R~\cite{wang2024spann3r}, and the flagship model VGG-T on Sintel. Notably, \nickname{}$^{\beta}$-W, using sliding-window attention~\cite{jiang2023mistral7b} for constant cache, exceeds \nickname{}$^{\beta}$ on Bonn and KITTI despite accessing only five past frames.


\subsection{3D Reconstruction}
\begin{table*}[t]
  \centering
  \caption{\textbf{3D Reconstruction Evaluation on 7-Scenes~\cite{Shotton_2013_CVPR}.} 
Despite operating in the streaming setting, our method delivers competitive performance, matching or even exceeding that of offline optimization-based methods that leverage global alignment.
\nickname{}$^{\beta}$-FA indicates adopting full attention in our trained model for 3D reconstruction.
}
  \footnotesize
  \setlength{\tabcolsep}{0.3em}
\resizebox{0.9\textwidth}{!}{
  \begin{tabularx}{\textwidth}{>{\centering\arraybackslash}m{2.6cm} >{\centering\arraybackslash}m{1.2cm} >{\centering\arraybackslash}X >{\centering\arraybackslash}X >{\centering\arraybackslash}X >{\centering\arraybackslash}X >{\centering\arraybackslash}X >{\centering\arraybackslash}X >{\centering\arraybackslash}m{0.9cm}}
    \toprule
    \multirow{3}{*}{\textbf{Method}} & \multirow{3}{*}{\textbf{Type}} 
    & \multicolumn{2}{c}{{Acc}$\downarrow$} & \multicolumn{2}{c}{{Comp}$\downarrow$} & \multicolumn{2}{c}{{NC}$\uparrow$} 
    & \multirow{3}{*}{\textbf{FPS}} \\
    \cmidrule(lr){3-4} \cmidrule(lr){5-6} \cmidrule(lr){7-8}
     & & Mean & Med. & Mean & Med. & Mean & Med. & \\
    \midrule
    VGG-T~\cite{wang2025vggt} & FA
      & \textbf{0.087} & \textbf{0.039} & \underline{0.091} & \textbf{0.039} & \textbf{0.787} & \textbf{0.890} & \underline{12.0} \\
    Fast3R~\cite{Yang_2025_Fast3R} & FA
      & 0.164 & 0.108 & {0.163} & 0.080 & 0.686 & 0.775 & \bf 30.92 \\
    DUSt3R-GA~\cite{wang2024dust3r} & Optim
      & {0.146} & {0.077} & 0.181 & {0.067} & {0.736} & {0.839} & 0.68 \\
    MASt3R-GA~\cite{mast3r_arxiv24} & Optim
      & 0.185 & 0.081 & {0.180} & 0.069 & 0.701 & 0.792 & 0.34 \\
    \monster{}-GA~\cite{zhang2024monst3r} & Optim
      & 0.248 & 0.185 & 0.266 & 0.167 & 0.672 & 0.759 & 0.39 \\
    \nickname{}$^{\beta}$-FA & Stream
      & \underline{0.091} &  \underline{0.043} &  \textbf{0.075} &  \underline{0.042} & \underline{0.769} & \underline{0.879} & \underline{12.0} \\
    \midrule
    Spann3R~\cite{wang2024spann3r} & Stream
      & 0.298 & 0.226 & 0.205 & 0.112 & 0.650 & 0.730 & 12.97 \\
    SLAM3R~\cite{slam3r} & Stream
      & 0.287 & 0.155 & 0.226 & 0.066 & 0.644 & 0.720 &  \bf 38.40 \\
    CUT3R~\cite{wang2025cut3r} & Stream
      &  \underline{0.126} & \underline{0.047} & \underline{0.154} & \underline{0.031} & \underline{0.727} & \underline{0.834} & 17.00 \\
    \nickname{}$^{\alpha}$ & Stream
      & 0.148 &  {0.077} &  {0.177} &  {0.058} & {0.700} & {0.801} & \underline{26.4} \\
    \nickname{}$^{\beta}$ & Stream
      & \bf{0.122} &  \bf{0.044} &  \textbf{0.110} &  \bf{0.038} & \bf{0.746} & \bf{0.856} & 20.12 \\
    \bottomrule
  \end{tabularx}
  }
  \label{tab:3d_recon}
  \vspace{-3mm}
\end{table*}

We also benchmark scene-level 3D reconstruction on the 7-scenes~\cite{Shotton_2013_CVPR} dataset and use accuracy (Acc), completion (Comp), and normal consistency (NC) metrics, following the convention of existing methods~\cite{wang2024spann3r,wang2025cut3r,wang2024dust3r}.
Following CUT3R, we assess the model's performance on image collections with minimal or no overlap by evaluating using sparsely sampled images, \ie, $3$ to $5$ frames per scene. The quantitative results are included in Tab.~\ref{tab:3d_recon}.
Our method achieves better performance compared to optimization-based methods and strong baselines including Spann3R, Fast3R, CUT3R, and SLAM3R. 
Compared to CUT3R, our method shows better performance with over $50\%$ times faster during the inference. 
While SLAM3R achieves the fastest inference, it yields noticeably lower reconstruction accuracy than our method. This performance gap can be partially attributed to SLAM3R being trained and evaluated at a lower input resolution of $224\times224$.
We also include \nickname{}$^{\beta}$-FA for comparison, which indicates replacing the causal attention in \nickname{}$^{\beta}$ into full attention (FA). 
Interestingly, \nickname{}$^{\beta}$-FA yields comparable performance compared to VGG-T and even better results on the completion metric. 
This highlights the effectiveness and generality of our proposed method.
The comparison results on NRGBD~\cite{Azinovic_2022_CVPR} benchmark is included in the supplement.

\subsection{Camera Pose Estimation}
\begin{table*}[t]
\caption{\textbf{Camera Pose Evaluation} on Sintel~\cite{sintel}, TUM-dynamic~\cite{tum-dynamics}, and ScanNet~\cite{dai2017scannet} datasets.
Our method achieves comparable performance with CUT3R on most benchmarks.
}
\centering
\footnotesize
\renewcommand{\arraystretch}{1.}
\renewcommand{\tabcolsep}{2.5pt}
\resizebox{0.975\textwidth}{!}{
\begin{tabular}{@{}clc|ccc|ccc|ccc@{}}
\toprule
& & & \multicolumn{3}{c}{\textbf{Sintel}} & \multicolumn{3}{c}{\textbf{TUM-dynamics}} & \multicolumn{3}{c}{\textbf{ScanNet}} \\ 
\cmidrule(lr){4-6} \cmidrule(lr){7-9} \cmidrule(lr){10-12}
{} & {\textbf{Method}}  & \textbf{Type} & {ATE $\downarrow$} & {RPE trans $\downarrow$} & {RPE rot $\downarrow$} & {ATE $\downarrow$} & {RPE trans $\downarrow$} & {RPE rot $\downarrow$} & {ATE $\downarrow$} & {RPE trans $\downarrow$} & {RPE rot $\downarrow$} \\ 
\midrule
& Particle-SfM~\cite{zhao2022particlesfm} & Optim & \underline{0.129} & 0.031 & \bf{0.535} & - & - & - & 0.136 & 0.023 & 0.836 \\ 
& Robust-CVD~\cite{kopf2021rcvd} & Optim & 0.360 & 0.154 & 3.443 & 0.153 & 0.026 & 3.528 & 0.227 & 0.064 & 7.374 \\ 
& CasualSAM~\cite{zhang2022structure} & Optim & 0.141 & \textbf{0.035} & \underline{0.615} & \underline{0.071} & \textbf{0.010} & 1.712 & 0.158 & 0.034 & 1.618 \\ 
& DUSt3R-GA~\cite{wang2024dust3r} & Optim & 0.417 & 0.250 & 5.796 & 0.083 & 0.017 & 3.567 & 0.081 & 0.028 & 0.784 \\  
& MASt3R-GA~\cite{mast3r_arxiv24} & Optim & 0.185 & 0.060 & 1.496 & \bf{0.038} & \underline{0.012} & \bf{0.448} & \underline{0.078} & \underline{0.020} & \bf{0.475} \\ 
& MonST3R-GA~\cite{zhang2024monst3r} & Optim & \bf{0.111} & \underline{0.044} & 0.869 & 0.098 & 0.019 & \underline{0.935} & \bf{0.077} & \bf{0.018} & \underline{0.529} \\ 
\midrule
& \duster{}~\cite{wang2024dust3r} & Onl & \underline{0.290} & 0.132 & 7.869 & 0.140 & 0.106 & 3.286 & 0.246 & 0.108 & 8.210 \\ 
& Spann3R~\cite{wang2024spann3r} & Onl & 0.329 & 0.110 & 4.471 & 0.056 & 0.021 & 0.591 & \underline{0.096} & 0.023 & \underline{0.661} \\ 
& CUT3R~\cite{wang2025cut3r} & Onl & \bf{0.213} & \bf{0.066} & \bf{0.621} & \underline{0.046} & \underline{0.015} & \underline{0.473} & 0.099 & \underline{0.022} & \bf{0.600}\\ 
& \nickname{}$^\beta$ & Onl & \bf{0.213} & \underline{0.076} & \underline{0.868} & \bf{0.026} & \bf{0.013} & \bf{0.330} & \bf{0.052} & \bf{0.021} & 0.850\\ 
\bottomrule
\end{tabular}
}
\label{tab:video_pose}
\end{table*}

Following CUT3R~\cite{wang2025cut3r}, we evaluate camera pose estimation accuracy on the Sintel~\cite{sintel}, TUM-dynamics~\cite{tum-dynamics}, and ScanNet~\cite{dai2017scannet} datasets.
Sintel and TUM-dynamics both feature substantial dynamic motion, 
posing significant challenges to conventional SfM and SLAM pipelines. 
We report Absolute Translation Error (ATE), Relative Translation Error (RPE$_{\text{trans}}$), and Relative Rotation Error (RPE$_{\text{rot}}$) after Sim(3) alignment with the ground truth, following the protocol in~\cite{teed2021droid, zhang2024monst3r, wang2025cut3r}. 
Our approach operates without requiring camera calibration, 
similar to the compared baselines~\cite{teed2021droid}. 
While many prior methods~\cite{kopf2021rcvd, zhang2022structure} 
address this via test-time optimization—jointly estimating 
intrinsics and dense depth for each sequence—we focus on purely online processing. 
Table~\ref{tab:video_pose} reports results for Online (Onl) and Optimization (Optim) categories, 
with DUSt3R~\cite{wang2024dust3r} included in the latter 
(aligning all frames to the first frame without global alignment). 
Although optimization-based systems still achieve the lowest errors overall, 
our method establishes the strongest performance among streaming approaches, 
and notably surpasses CUT3R~\cite{wang2025cut3r} on both TUM-dynamics and ScanNet, 
demonstrating particular robustness in dynamic environments.

\subsection{Ablation on the Effectiveness of the Proposed Architecture}
Here, we conduct detailed ablation analysis on \nickname{} to demonstrate the effectiveness of its designs. Due to the extensive computational resources required to train the model, we only train the ablation models on $224\times224$ resolution images. All the datasets are included to train the models.
Note that for a fair comparison, we initialize all the models below using the pre-trained MASt3R~\cite{mast3r_arxiv24} checkpoints and train the models using the same hyper-parameters and compute resources.
\begin{figure}[htbp]
    \centering
    \begin{subcaptionbox}{Overall Training Curve\label{fig:abla:training_loss}}[0.32\textwidth]
        {\includegraphics[width=\linewidth]{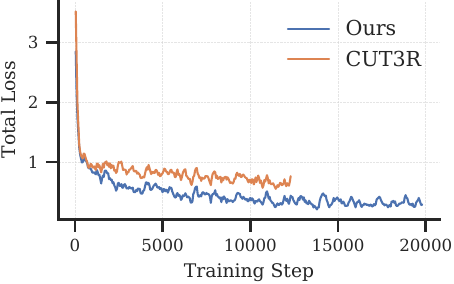}}
    \end{subcaptionbox}
    \hfill
    \begin{subcaptionbox}{Training Curve of Local Branch \label{fig:abla:local_loss}}[0.32\textwidth]
        {\includegraphics[width=\linewidth]{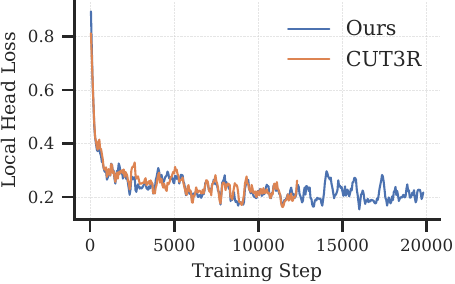}}
    \end{subcaptionbox}
    \hfill
    \begin{subcaptionbox}{Training Curve of Global Branch\label{fig:abla:global_loss}}[0.32\textwidth]
        {\includegraphics[width=\linewidth]{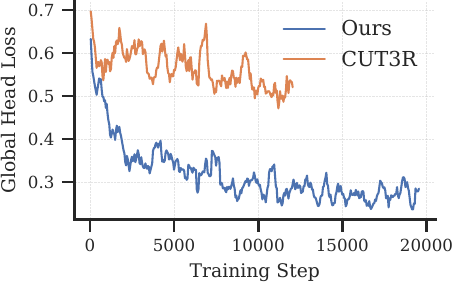}}
    \end{subcaptionbox}
    \caption{Ablation of our proposed \nickname{}. Compared to RNN-based architecture~\cite{wang2025cut3r}, our decoder-only network yields better convergence with faster training speed in the 3D point map prediction task, especially in the global branch.}
    \label{fig:ablation_with_rnn}
\end{figure}

We demonstrate the effectiveness of decoder-only transformer against RNN design in the sequential 3D pointmap prediction. The main baseline is CUT3R~\cite{wang2025cut3r}, which leverages the RNN design to achieve this. 
For a fair comparison, we re-train CUT3R and our method using the same dataset and pre-trained model weights initialization.
We include the training curve in Fig.~\ref{fig:abla:training_loss}, where both models are trained with the same hyperparameters and compute resources.
As can be observed, \nickname{} converges faster compared to CUT3R and performs $60\%$ more training steps within the given time.
This may sound counterintuitive since \nickname{} is attending to a longer context against CUT3R's constant \textit{state} memory.
However, since CUT3R architecture requires a \textit{state-update} operation after each \textit{state-readout} interaction, while \nickname{} directly attends to cached features of existing observations.

We also notice in Fig.~\ref{fig:abla:local_loss} that the convergence of $\headself$ is similar among the two architectures, while for $\headcross$, our proposed architecture shows noticeably faster convergence speed, as shown in Fig.~\ref{fig:abla:global_loss}.
This demonstrates that using a single \textit{state} makes the model harder to register incoming frames due to the limited memory capacity.

Quantitatively, we benchmark the ablation models on both the video depth estimation in Tab.~\ref{tab:abla:video_depth} and 3D reconstruction in Tab.~\ref{tab:abla_3d_recon}, which evaluates the $\headself$ and $\headcross$ correspondingly. 
For a fair comparison, we evaluate the checkpoints trained for the same number of iterations.
As can be observed, our proposed architecture consistently achieves better performance on both tasks.

\begin{table*}[t]
  \centering
  \caption{{
\textbf{Ablation on Video Depth Estimation}. 
When evaluating the checkpoint trained for the same number of iterations, our proposed architecture consistently achieves better performance against RNN-based CUT3R on the video depth estimation task.
}}
  \footnotesize
  \setlength{\tabcolsep}{0.5em}
  \resizebox{0.85\textwidth}{!}{
  \begin{tabularx}{\textwidth}{>{\centering\arraybackslash}m{1.5cm} 
                                 >{\centering\arraybackslash}X 
                                 >{\centering\arraybackslash}X 
                                 >{\centering\arraybackslash}X 
                                 >{\centering\arraybackslash}X 
                                 >{\centering\arraybackslash}X 
                                 >{\centering\arraybackslash}X}
\toprule
\multirow{3}{*}{\textbf{Method}} 
  & \multicolumn{2}{c}{\textbf{Sintel}} 
  & \multicolumn{2}{c}{\textbf{BONN}} 
  & \multicolumn{2}{c}{\textbf{KITTI}} \\
\cmidrule(lr){2-3} \cmidrule(lr){4-5} \cmidrule(lr){6-7}
  & Abs Rel $\downarrow$ & $\delta$\textless $1.25\uparrow$ 
  & Abs Rel $\downarrow$ & $\delta$\textless $1.25\uparrow$ 
  & Abs Rel $\downarrow$ & $\delta$\textless $1.25\uparrow$ \\
\midrule
CUT3R & 0.598 & 40.7 & 0.102 & 90.7 & 0.157 & 77.4 \\
\nickname{}$^{\alpha}$ & \textbf{0.535} & \textbf{47.0} & \textbf{0.083} & \textbf{94.2} & \textbf{0.141} & \textbf{81.8} \\
\bottomrule
\end{tabularx}
}
\label{tab:abla:video_depth}
\end{table*}
\begin{table*}[t]
  \centering
  \caption{
  \textbf{Ablation on 3D Reconstruction on 7-Scenes}.
  Our proposed architecture consistently achieves better performance against RNN-based CUT3R on the 3D reconstruction task when trained under the same configurations. Note that our architecture is trained even faster.
  }
  \footnotesize
  \setlength{\tabcolsep}{0.5em}
  \resizebox{0.85\textwidth}{!}{
  \begin{tabularx}{\textwidth}{>{\centering\arraybackslash}m{1.5cm} 
                                 >{\centering\arraybackslash}X 
                                 >{\centering\arraybackslash}X 
                                 >{\centering\arraybackslash}X 
                                 >{\centering\arraybackslash}X 
                                 >{\centering\arraybackslash}X 
                                 >{\centering\arraybackslash}X}
    \toprule
    \multirow{3}{*}{\textbf{Method}} 
      & \multicolumn{2}{c}{{Acc}$\downarrow$} 
      & \multicolumn{2}{c}{{Comp}$\downarrow$} 
      & \multicolumn{2}{c}{{NC}$\uparrow$} \\
    \cmidrule(lr){2-3} \cmidrule(lr){4-5} \cmidrule(lr){6-7}
      & Mean & Med. & Mean & Med. & Mean & Med. \\
    \midrule
    CUT3R
      & 0.480 & 0.365 &  0.330 &  0.148 & {0.555} & {0.583} \\
    \nickname{}$^{\alpha}$
      & \bf 0.328 &  \bf {0.261} &  \bf {0.255} & \bf  {0.095} & \bf {0.605} & \bf {0.659} \\
    \bottomrule
  \end{tabularx}
  }
  \label{tab:abla_3d_recon}
\end{table*}

\section{Conclusion and Discussions}
\label{sec:conclusion_discussion_limitation}
We have introduced \nickname{}, a decoder-only transformer framework for dense 3D reconstruction from unstructured or streaming image inputs. By reformulating reconstruction as a sequential registration task with causal attention, \nickname{} overcomes the scalability bottlenecks of prior work and aligns naturally with LLM-style training and inference pipelines. Our design allows efficient integration of geometric context across frames, supports dual-coordinate pointmap prediction, and generalizes to novel-view synthesis over large-scale scenes without the need for global post-processing.
Through extensive experiments across standard benchmarks, we show that \nickname{} achieves competitive or superior performance in the monocular/video-depth estimation and 3D reconstruction tasks, with significantly improved inference efficiency. By bridging geometric learning with scalable sequence modeling, we hope this work paves the way toward more general-purpose, real-time 3D understanding systems.

Our method comes with some limitations.
First, the na\"{i}ve causal modeling naturally suffers from error accumulation and drifting~\cite{zhang2025framepack}. 
Some inference strategies can be proposed to alleviate this issue.
Second, currently \nickname{} is still a regression model with deterministic outputs. Extending it further into an autoregressive generative model~\cite{chen2025diffusionforcing,zhang2025framepack} shall further unlock a series of downstream applications.
Finally, since \nickname{} follows a similar design of modern LLMs, more training techniques like MLA~\cite{deepseekai2024deepseekv2strongeconomicalefficient} can be introduced to further boost the training efficiency and performance.



\clearpage
{
\small
\bibliographystyle{unsrt}
\bibliography{bibs/cvpr24}
}

\newpage
\appendix

\appendix
\section{Dataset Details}
We train our model on $29$ datasets that contains a diverse range of scene types, including static and dynamic scene and objects. Specifically, we mainly follow the data splits of CUT3R~\cite{wang2025cut3r}, and the main $15$ datasets with highest sampling ratio are:
Co3Dv2~\cite{reizenstein21co3d}, 
ScanNet++~\cite{yeshwanth2023scannet++},
ScanNet~\cite{dai2017scannet}, 
HyperSim~\cite{hypersim}, 
Dynamic Replica~\cite{karaev2023dynamicstereo}, 
DL3DV~\cite{ling2024dl3dv}, 
BlendedMVS~\cite{yao2020blendedmvs}, 
Aria Synthetic Environments~\cite{Pan_2023_ariadigitaltwin},
TartanAir~\cite{tartanair2020iros}, 
MapFree~\cite{arnold2022mapfree}, 
MegaDepth~\cite{Li2018MegaDepthLS}, 
WildRGBD~\cite{xia2024rgbdobjectswildscaling},
Waymo~\cite{waymo},
Bedlam~\cite{Black_CVPR_2023},
and ARKitScenes~\cite{dehghan2021arkitscenes}.
We do not include 3D Ken Burns~\cite{kenburnsdataset}, IRS~\cite{wang2021irslargenaturalisticindoor}, and SmartPortraits~\cite{smartportraits} for training since these datasets are either single view or fail to download successfully.
We adapt the official scripts provided by CUT3R~\cite{wang2025cut3r}, DUSt3R~\cite{wang2024dust3r}, and Spann3R~\cite{wang2024spann3r} for dataset processing.
For training \nickname{}$^\beta$, we remove all the single-view datasets as in VGG-T, leaving $19$ datasets for training. We did not find performance degradation when removing the single-view datasets.
Please refer to the Tab. 6 of the CUT3R for more dataset details.

\section{More Implementation details}
\heading{More Training Details}
Our method conducts end-to-end training on all datasets on a hybrid of $12$ different resolutions, ranging from $224\times224$ to $512\times384$. Data augmentation side, we perform sequence-level color jittering by applying the same color jitter across all frames in a sequence.

\heading{Network Architecture Details}
We follow DUSt3R and use the CroCoNet~\cite{croco_v2} pre-trained ViT for the encoder and decoder design. We directly use the DPT~\cite{dpt} head for $\headcross$ and $\headself$ implementation. We apply RoPE to the query and key feature before each attention operation for the ViT encoder, but ignore it for the ViT decoder to generalize to an arbitrary number of input views. For ablation studies, we train our model on the same datasets but at resolution $224\times224$.

For the sliding window attention version \nickname{}$^\beta$-W[5], we always include the tokens of the first frame to keep the canonical coordinate space unchanged. We set window size W$=5$ since it trades off performance and speed, and other window size also stably works.
For the full attention version \nickname{}$^\beta$-FA, we directly use the causally trained model \nickname{}$^\beta$ and remove the causal mask in the $\mathrm{SelfAttn}$. This is similar to the ``revisit'' operation in CUT3R.

\section{More Comparisons}
\begin{table*}[t]
\centering
\caption{\small{
\textbf{Video Depth Evaluation}. We report scale\&shift-invariant depth, scale-invariant depth and metric depth accuracy on Sintel, Bonn, and KITTI datasets. Methods requiring global alignment are marked ``GA'', while ``Optim'' and ``Stream'' indicate Optimzation-based and Streamne methods, respectively. We also report the FPS on KITTI dataset using 512$\times$ 144 image resolution for all methods, except \spanner{} which Stream supports 224$\times$224 inputs.
}
}
\renewcommand{\arraystretch}{1.02}
\renewcommand{\tabcolsep}{1.5pt}
\resizebox{0.99\textwidth}{!}{
\begin{tabular}{@{}llc>{\centering\arraybackslash}p{1.5cm}>{\centering\arraybackslash}p{1.5cm}|>{\centering\arraybackslash}p{1.5cm}>{\centering\arraybackslash}p{1.5cm}|>{\centering\arraybackslash}p{1.5cm}>{\centering\arraybackslash}p{1.5cm}|>{\centering\arraybackslash}p{1.2cm}@{}}
\toprule
 &  &  & \multicolumn{2}{c}{\textbf{Sintel}} & \multicolumn{2}{c}{\textbf{BONN}} & \multicolumn{2}{c}{\textbf{KITTI}} & \\ 
\cmidrule(lr){4-5} \cmidrule(lr){6-7} \cmidrule(lr){8-9}
\textbf{Alignment} & \textbf{Method} & \textbf{Type} & {Abs Rel $\downarrow$} & {$\delta$\textless $1.25\uparrow$} & {Abs Rel $\downarrow$} & {$\delta$\textless $1.25\uparrow$} & {Abs Rel $\downarrow$} & {$\delta$ \textless $1.25\uparrow$} & \textbf{FPS} \\ 
\midrule
\multirow{13}{*}{\begin{minipage}{3cm}Per-sequence \\
scale  \& shift\end{minipage}} 
 & Marigold~\cite{ke2024repurposing} & Stream & 0.532 & {51.5} & {0.091} & {93.1} & {0.149} & {79.6} & $<$0.1 \\
  & Depth-Anything-V2~\cite{yang2024depthv2} & Stream & 0.367 & {55.4} & {0.106} & {92.1} &{0.140} & {80.4} & 3.13 \\
   & NVDS~\cite{NVDS} & Stream & 0.408 & {48.3} & {0.167} & {76.6} & {0.253} & {58.8} & - \\
    & ChronoDepth~\cite{shao2024learningtemporallyconsistentvideo} & Stream & 0.687 & {48.6} & {0.100} & {91.1} & {0.167} &{75.9} & 1.89 \\
     & DepthCrafter~\cite{hu2024-DepthCrafter} & Stream & \textbf{0.292} & \textbf{{69.7}} & {0.075} & \textbf{{97.1}} & {{0.110}} & {{88.1}} & 0.97 \\
      & Robust-CVD~\cite{kopf2021rcvd} & Stream & 0.703 & {47.8} & {-} & {-} & - & - & - \\
      & CasualSAM~\cite{zhang2022structure} & Optim & 0.387 & {54.7} & {0.169} & {73.7} & {0.246} & 62.2 & - \\
& DUSt3R-GA~\cite{wang2024dust3r} & Optim & 0.531 & {51.2} & {0.156} & {83.1} & {0.135} & {81.8} & 0.76 \\
& MASt3R-GA~\cite{mast3r_arxiv24} & Optim & \underline{0.327} & \underline{59.4} & {0.167} & {78.5} & {0.137} & {83.6} & 0.31 \\

& MonST3R-GA~\cite{zhang2024monst3r} & Optim & {0.333} & {59.0} & \textbf{{0.066}} & \underline{96.4} & {0.157} & {73.8} & 0.35 \\
& Spann3R~\cite{wang2024spann3r} & Stream & 0.508 & {50.8} & {0.157} & {82.1} & {0.207} & {73.0} &13.55 \\

& CUT3R~\cite{wang2025cut3r} & Stream & 0.540  & 55.7 & {0.074} & {94.5} & {0.106} & \underline{88.7} &  16.58 \\
& \nickname{} & Stream & 0.356  & 58.6 & \underline{0.068} & {95.7} & \bf{0.099} & \bf{91.0} &  \bf 23.48 \\

\midrule
\multirow{6}{*}{\begin{minipage}{3cm}Per-sequence scale\end{minipage}} & DUSt3R-GA~\cite{wang2024dust3r} & Optim & 0.656 & {45.2} & {0.155} & {83.3} & \underline{0.144} & {81.3} & 0.76 \\
& MASt3R-GA~\cite{mast3r_arxiv24} & Optim & 0.641 & {43.9} & {0.252} & {70.1} & {0.183} & {74.5} & 0.31 \\

& MonST3R-GA~\cite{zhang2024monst3r} & Optim & \textbf{0.378} & \textbf{55.8} & \textbf{0.067} & \textbf{96.3} & {0.168} & {74.4} & 0.35 \\
& Spann3R~\cite{wang2024spann3r} & Stream & 0.622 & {42.6} & {0.144} & {81.3} & {0.198} & {73.7} &13.55 \\
& Fast3R~\cite{Yang_2025_Fast3R} & FA & 0.653 & {44.9} & {0.193} & {77.5} & {0.140} & {83.4} & \bf 47.23 \\
& CUT3R~\cite{wang2025cut3r} & Stream & \underline{0.421}  & {47.9} & {0.078} & {93.7} & \underline{0.118} & \underline{88.1} &  16.58 \\
& \nickname{}$^\alpha$ & Stream & {0.478}  & \underline{51.1} & \underline{0.075} & \underline{94.1} & \textbf{0.116} & \textbf{89.6} &  \underline{23.48} \\

\midrule
 \multirow{2}{*}{\begin{minipage}{3cm}Metric scale\end{minipage}} & MASt3R-GA~\cite{mast3r_arxiv24} & Optim & \textbf{1.022}  & 14.3 & 0.272 & 70.6 & 0.467 & 15.2  & 0.31 \\  
& CUT3R & Stream & \underline{1.029} & \textbf{23.8} & \underline{0.103} & \underline{88.5}& \textbf{0.122} & \textbf{85.5}  &  16.58 \\
& \nickname{}$^\alpha$ & Stream & {1.041} & \underline{21.0} & \textbf{0.084} & \textbf{94.4}& \underline{0.234} & \underline{57.6}  & \bf 23.48 \\
\bottomrule
\end{tabular}
}
\label{tab:video_depth_supp}
\end{table*}
\heading{Video Depth Estimation}
We further expand the video depth comparison in the main paper and include a wider range of baseline methods, including single-frame depth methods (Marigold~\cite{ke2024repurposing} and DepthAnything-V2~\cite{yang2024depthv2}), video depth approaches (NVDS~\cite{NVDS}, DepthCrafter~\cite{hu2024-DepthCrafter}, and ChronoDepth~\cite{shao2024learningtemporallyconsistentvideo}), and recent joint depth-and-pose estimation methods such as Robust-CVD~\cite{barsan2018robust}, CausalSAM~\cite{zhang2022structure}, DUSt3R~\cite{wang2024dust3r}, MASt3R~\cite{mast3r_arxiv24}, MonST3R~\cite{zhang2024monst3r}, and Spann3R~\cite{wang2024spann3r}. 
Extended results are shown in Tab.~\ref{tab:video_depth_supp}. \nickname{} consistently outperforms its RNN-based counterpart CUT3R under the per-sequence scale $\&$ shift setting, and even achieves state-of-the-art performance on the KITTI dataset—while also being the fastest in terms of FPS.

\begin{table*}[t]
  \centering
  \caption{\textbf{3D Reconstruction Comparison on NRGBD~\cite{Azinovic_2022_CVPR}.} 
Our proposed method consistently achieves superior performance compared to optimization-based (Optim), streaming-based (Stream), and even full attention (FA) methods.}
  \footnotesize
  \setlength{\tabcolsep}{0.3em}
\resizebox{0.9\textwidth}{!}{
  \begin{tabularx}{\textwidth}{>{\centering\arraybackslash}m{3.2cm} >{\centering\arraybackslash}m{1.2cm} >{\centering\arraybackslash}X >{\centering\arraybackslash}X >{\centering\arraybackslash}X >{\centering\arraybackslash}X >{\centering\arraybackslash}X >{\centering\arraybackslash}X}
    \toprule
    \multirow{2}{*}{\textbf{Method}} & \multirow{2}{*}{\textbf{Type}} 
    & \multicolumn{2}{c}{{Acc}$\downarrow$} & \multicolumn{2}{c}{{Comp}$\downarrow$} & \multicolumn{2}{c}{{NC}$\uparrow$} \\
    \cmidrule(lr){3-4} \cmidrule(lr){5-6} \cmidrule(lr){7-8}
     & & Mean & Med. & Mean & Med. & Mean & Med. \\
    \midrule
    VGG-T~\cite{wang2025vggt} & FA
      & \bf 0.073 & \bf 0.018 & \bf 0.077 & \bf 0.021 & \bf 0.910 & \bf 0.990 \\
    DUSt3R-GA~\cite{wang2024dust3r} & Optim
      & 0.144 & \underline{0.019} & 0.154 & \underline{0.018} & \underline{0.870} & \underline{0.982} \\
    MASt3R-GA~\cite{mast3r_arxiv24} & Optim
      & \underline{0.085} & 0.033 & \underline{0.063} & 0.028 & 0.794 & 0.928 \\
    \monster{}-GA~\cite{zhang2024monst3r} & Optim
      & 0.272 & 0.114 & 0.287 & 0.110 & 0.758 & 0.843 \\
    \midrule
    Spann3R~\cite{wang2024spann3r} & Stream
      & 0.416 & 0.323 & 0.417 & 0.285 & 0.684 & 0.789 \\
    CUT3R~\cite{wang2025cut3r} & Stream
      & 0.099 & \underline{0.031} & 0.076 & \underline{0.026} & 0.837 & 0.971 \\
    StreamVGGT~\cite{streamVGGT} & Stream
      & \underline{0.084} & 0.044 & \underline{0.074} & 0.041 & \underline{0.861} & \underline{0.986} \\
    VGG-T [Streaming]~\cite{wang2025vggt} & Stream
      & 0.219 & 0.102 & 0.212 & 0.105 & 0.797 & 0.936 \\
    \nickname{}$^{\beta}$ & Stream
      & \bf 0.057 & \bf 0.014 & \bf 0.028 & \bf 0.013 & \bf 0.910 & \bf 0.993 \\
    \bottomrule
  \end{tabularx}
}
  \label{tab:3d_recon_nrbgd}
  \vspace{-3mm}
\end{table*}

\heading{3D Reconstruction on NRGBD}
We further include the comparison on NRGBD benchmark~\cite{Azinovic_2022_CVPR} in Tab.~\ref{tab:3d_recon_nrbgd}.
Here, we also include the comparison with a concurrent work StreamVGGT~\cite{streamVGGT}, which fine-tunes VGG-T into streaming version similar to our method. 
We also include VGG-T[streaming], which indicates using VGG-T in the streaming setting by replace the full attention in VGG-T into the causal attention.
As can be seen, our method clearly outperforms all optimization-based and online methods, including the official VGG-T model. 
Direct use of VGG-T in the streaming setting substantially degrades performance, underscoring the need for fine-tuning under causal constraints.




\end{document}